\documentclass[letterpaper, 10 pt, conference]{ieeeconf}
\IEEEoverridecommandlockouts

\usepackage{amsmath}
\DeclareMathOperator*{\argmax}{arg\,max}

\usepackage{cite}
\usepackage{amssymb}
\usepackage[ruled,linesnumbered,noline]{algorithm2e}
\usepackage{float}
\usepackage{textcomp}
\usepackage{bm}

\newtheorem{definition}{Definition}

\usepackage{makecell}

\usepackage{graphicx}
\usepackage{tikz}
\usepackage{float}
\usetikzlibrary{arrows}
\usetikzlibrary{arrows.meta}
\usetikzlibrary{arrows,positioning} 
\usetikzlibrary{shapes}
\usetikzlibrary{calc}
\usetikzlibrary{fadings}
\usetikzlibrary{decorations.pathreplacing}
\usetikzlibrary{decorations.text}

\usepackage{hyperref}
\usepackage{xcolor}
\hypersetup{
    colorlinks,
    linkcolor={black},
    citecolor={blue!50!black},
    urlcolor={blue!80!black}
}
\usepackage{subfigure}

\usepackage{amsfonts}

\usepackage{scalerel}
\def\msquare{\mathord{\scalerel*{\Box}{gt}}}
\def\mdiamond{\mathord{\scalerel*{\Diamond}{gt}}}
\makeatletter \providecommand\dotdiamond{\mathpalette\@barred\mdiamond} \def\@barred#1#2{\ooalign{\hfil$#1\cdot$\hfil\cr\hfil$#1#2$\hfil\cr}}  \makeatother
\makeatletter \providecommand\dotbox{\mathpalette\@burrow\msquare} \def\@burrow#1#2{\ooalign{\hfil$#1\cdot$\hfil\cr\hfil$#1#2$\hfil\cr}}  \makeatother

\usepackage[top=54pt, left=54pt, right=54pt, bottom=54pt]{geometry}
\begin{document}

\title{\textbf{\huge Bipedal Safe Navigation over Uncertain Rough Terrain: Unifying Terrain Mapping and Locomotion Stability}\vspace*{3pt}}

\author{Kasidit~Muenprasitivej*, Jesse~Jiang*, Abdulaziz~Shamsah*, Samuel~Coogan, and Ye~Zhao \thanks{*Equally contributed authors.}
\thanks{This work was supported in part by a National Science Foundation Graduate Research Fellowship under grant \#DGE-2039655.}
\thanks{The authors are with the Institute of Robotics and Intelligent Machines, Georgia Institute of Technology, Atlanta, GA 30332, USA (email: \{kmuenpra3, jjiang, ashamsah3, sam.coogan\}@gatech.edu, ye.zhao@me.gatech.edu). 
}}
\author{Kasidit~Muenprasitivej*$^{1}$, Jesse~Jiang*$^{2}$, Abdulaziz~Shamsah*$^{3,4}$, Samuel~Coogan$^{2,5}$, and Ye~Zhao$^{3}$ \thanks{*Equally contributed authors.}
\thanks{This work was supported in part by a National Science Foundation Graduate Research Fellowship under grant \#DGE-2039655.}
\thanks{$^{1}$Daniel Guggenheim School of Aerospace Engineering, Georgia Institute of Technology, Atlanta, GA 30332 USA (e-mail: kmuenpra3@gatech.edu)}
\thanks{$^{2}$School of Electrical and Computer Engineering, Georgia Institute of Technology, Atlanta, GA 30332 USA (e-mail: jjiang@gatech.edu, sam.coogan@gatech.edu).}
\thanks{$^{3}$George W. Woodruff School of Mechanical Engineering, Georgia Institute of Technology, Atlanta, GA, 30332 USA (e-mail: ashamsah3@gatech.edu, ye.zhao@me.gatech.edu).}
\thanks{$^{4}$Mechanical Engineering Department, College of Engineering and Petroleum, Kuwait University, PO Box 5969, Safat, 13060, Kuwait}
\thanks{$^{5}$School of Civil and Environmental Engineering, Georgia Institute of Technology, Atlanta, GA 30332 USA}
}

\maketitle

\thispagestyle{empty}
\begin{abstract}
We study the problem of bipedal robot navigation in complex environments with uncertain and rough terrain. In particular, we consider a scenario in which the robot is expected to reach a desired goal location by traversing an environment with uncertain terrain elevation. Such terrain uncertainties induce not only untraversable regions but also robot motion perturbations. Thus, the problems of terrain mapping and locomotion stability are intertwined. We evaluate three different kernels, including two non-stationary kernels, for Gaussian process (GP) regression to learn the terrain elevation. We also learn the motion deviation resulting from both the terrain as well as the discrepancy between the reduced-order Prismatic Inverted Pendulum Model used for planning and the full-order locomotion dynamics. We propose a hierarchical locomotion-dynamics-aware sampling-based navigation planner. The global navigation planner plans a series of local waypoints to reach the desired goal locations while respecting locomotion stability constraints. Then, a local navigation planner is used to generate a sequence of dynamically feasible footsteps to reach local waypoints. We develop a novel trajectory evaluation metric to minimize motion deviation and maximize information gain of the terrain elevation map.  
\renewcommand{\thefootnote}{\roman{footnote}}
We evaluate the efficacy of our planning framework on Digit bipedal robot simulation in MuJoCo.\footnotemark[1]\footnotetext[1]{Videos of the simulated experiments in Mujoco can be found at \url{https://youtu.be/27hOcBNKAvU?si=dp3GSGHndtQ7ZMaC}}
\end{abstract}

\begin{keywords}
bipedal robot, terrain mapping, locomotion stability, hierarchical sampling-based navigation, uncertainty modelling, path evaluation, Gaussian process learning
\end{keywords}

\section{Introduction}
Legged robots show great promise for navigation tasks in environments with difficult-to-traverse or unknown terrain. As opposed to wheeled mobile robots, legged robots have the superior capability of traversing through irregular terrains by taking discrete footsteps \cite{torres2022legged, gibson2022terrain,huang2023efficient}. However, highly varying and uncertain terrain profiles often induce tracking errors when executing bipedal motion plans or even pose a high risk in locomotion failures (\textit{i.e.}, falling)\cite{DataS2S_Dai,krishna2022linear, wu2023infer}. Thus, navigation through complex and uncertain terrain requires collecting terrain data online to build a realistic terrain map and improve locomotion performance accordingly. On the other hand, the complex dynamics inherent to bipedal locomotion complicate the problem of designing navigation plans to sample the environment. Thus, the objectives of locomotion stability (\textit{i.e.}, minimizing motion deviation from the desired stable trajectory in this study) and accurate environmental sampling are coupled, increasing the complexity of the entire navigation problem.

\begin{figure}[t]
    \centering
    \includegraphics[width=0.85\linewidth]{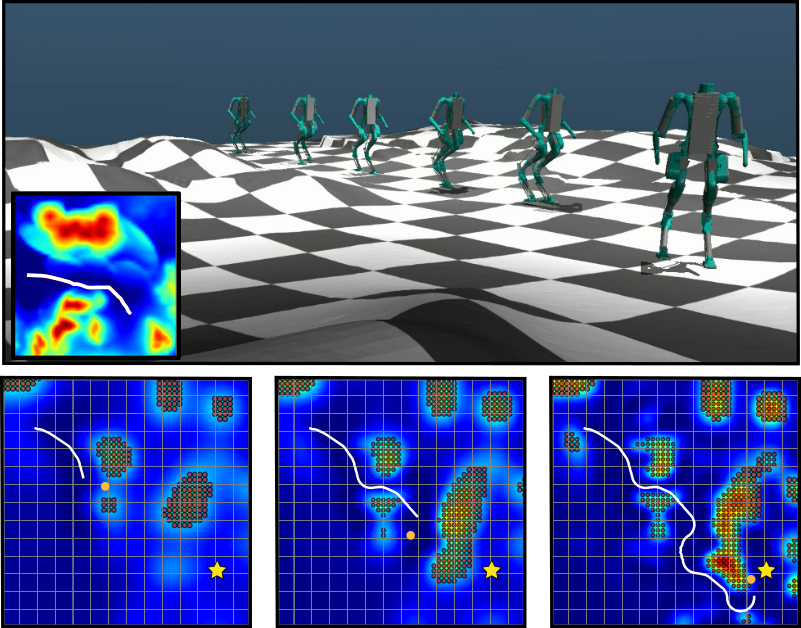}
    \caption{(Top) The bipedal robot Digit navigates through an environment with rough terrain in our MuJoCo simulation. (Bottom) Snapshots of the trajectory of the bipedal robot at various time instants as it navigates towards the goal (yellow star). The white line depicts the traversed trajectory, and the orange dot is the current targeted local waypoint.}
    \label{fig:highlight}
    \vspace{-0.2in}
\end{figure}

In this work, we propose a hierarchical planning strategy for bipedal robots which satisfies high-level global navigation objectives while maintaining dynamic feasibility of the generated trajectories in the local navigation planner. Additionally, we use Gaussian processes (GPs) with three different kernels to learn unknown terrain elevation. We also learn motion perturbation resulting from both terrain and model errors. Our planner is designed to incorporate the GP predictions in order to online improve  the feasibility of reaching the desired goal. An example run of our planner is shown in Figure \ref{fig:highlight}.
\subsection{Related Works}
The RRT family of algorithms is commonly used in concert with GPs for robotic motion planning problems in uncertain environments. The study in \cite{yang2013gaussian} considers an aerial vehicle navigation problem and uses RRT to navigate around collision regions modeled using a GP. The work \cite{barbosa2021risk} uses RRT* to enable a mobile robot to avoid hazardous regions which are learned and updated online using a GP. For an information-gathering objective, the paper \cite{viseras2019robotic} learns optimal points to sample using a GP model of the environment and uses RRT* to plan information-gain-maximizing trajectories. Finally, the work \cite{jian2023path} trains a GP model of terrain elevation using both external perception and proprioception sensors and then proposes an RRT* variant to plan a safe trajectory despite environmental occlusion.

\begin{figure}[t]
    \centering
    \includegraphics[width=0.8\linewidth]{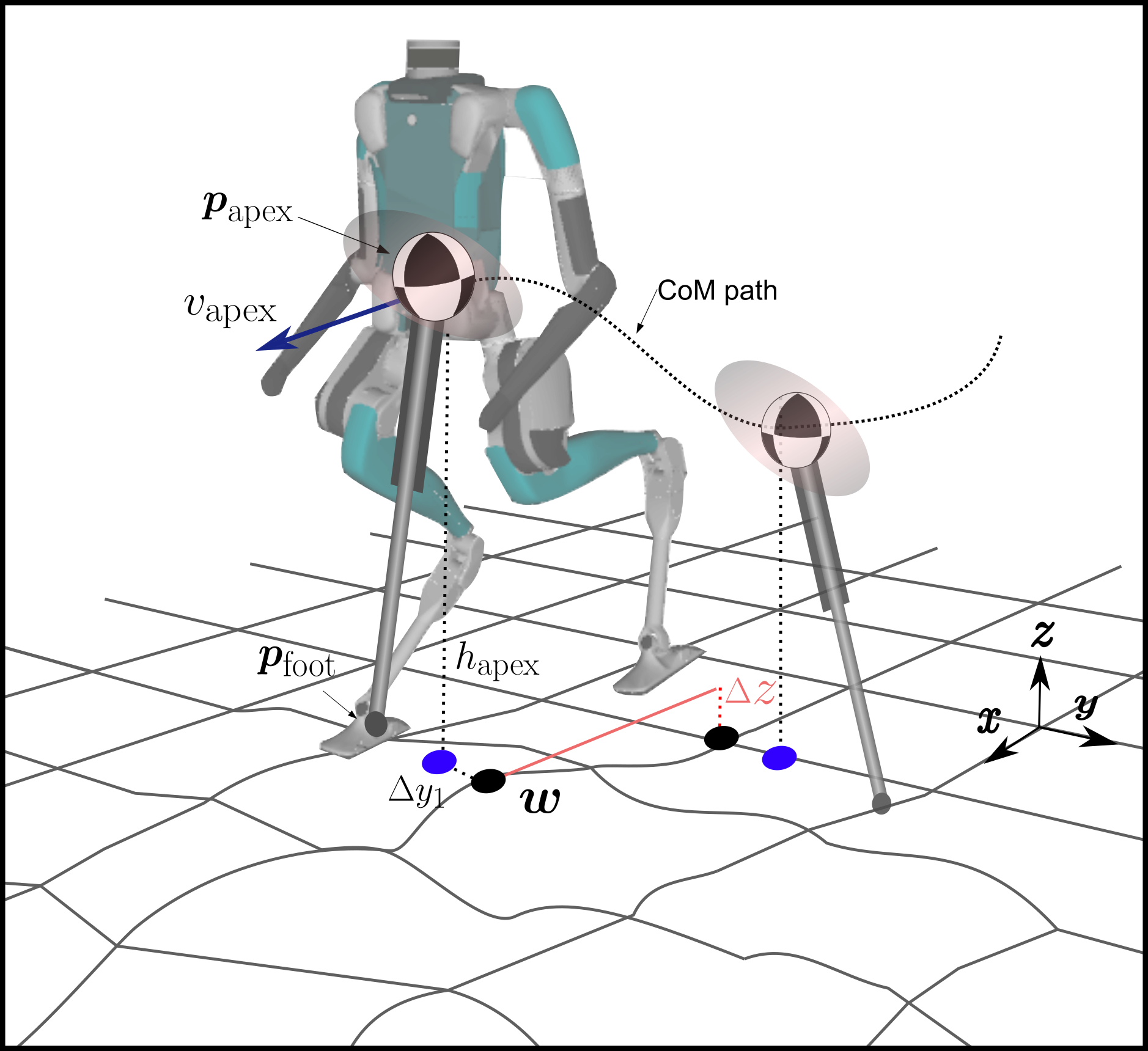}
    \caption{Prismatic Inverted Pendulum Model (PIPM) model for our robot Digit for traversing over uncertain and uneven terrain.}
    \label{fig:notation}
    \vspace{-0.2in}
\end{figure}

The problem of bipedal robot navigation in rough terrain has not yet been widely explored. The studies in  \cite{kanoulas2018footstep, bertrand2020detecting} propose methods for identifying stable footstep sequences using sensor data which allow bipedal robots to traverse over uneven terrain. The focus of these works is on finding stable local trajectories rather than long-run trajectories to reach global goals. The authors in~\cite{gibson2022terrain} propose a terrain-adaptive bipedal locomotion controller that uses a piecewise linear terrain approximation for computing foot placements. The work in~\cite{huang2023efficient} proposes an omnidirectional control Lyapunov function (CLF) as a controller for a bipedal robot navigating on undulating terrain and integrates the CLF into a RRT* planner. The relative elevation, slope, and friction are used as a traversability cost in the planner. In this work, an elevation map is constructed online using sensor data, but is not otherwise learned. Additionally, the omnidirectional nature of the planner relies on special behaviors such as turn-in-place and lateral stepping for bipedal locomotion feasibility.

From a terrain mapping perspective, GPs have been widely used in the literature to quantify terrain uncertainties and learn complex terrain maps. The work in \cite{chen2022ak} proposes a novel nonstationary kernel which is used to efficiently learn unknown environmental features by prioritizing areas with higher variation for exploration. This kernel is also designed to accurately model rapidly varying terrain by using a mixture model of base kernels and datasets. Another approach is the neural network kernel explained in \cite{vasudevan2009gaussian}, which designs a GP approximating a simple neural network while retaining the information-theoretic learning guarantees of Gaussian process theory. The nonstationary nature of this kernel is well-suited for learning discontinuous data. Another aspect of this work is the use of KD-trees to reduce the dataset size for GP calls, increasing computational tractability. 

Leveraging GP approaches to map the terrain has gained increasing attention in the locomotion community. The work in~\cite{plagemann2009bayesian} implements a locally adaptive GP for terrain mapping in a legged navigation problem for the Boston Dynamics LittleDog quadruped. The proposed local GP framework seeks to balance the fidelity of the learned model with the computational tractability of training the GP. The authors of \cite{seyde2019locmotion} use GPs to evaluate candidate trajectories for a hopping robot locomotion planning problem. The work~\cite{homberger2019support} learns a GP-based terrain map from sensor data and uses the GP model to design foothold placements for the ETH ANYmal quadrupedal robot. However, learning terrain uncertainty via GP models for bipedal robot navigation has not been explored, to the best of the authors' knowledge. The inherent stability-critical, complex robot dynamics make the terrain learning and navigation problem more challenging.

\subsection{Contributions}
We propose a novel hierarchical planning framework for bipedal robot locomotion with high-level navigation tasks that generates dynamically feasible locomotion trajectories while simulataneously learning unknown terrain features. Our specific contributions are as follows.
\begin{itemize}
    \item We propose a hierarchical locomotion-dynamics-aware planner based on RRT* which enables computationally efficient bipedal navigation while explicitly considering dynamical feasibility of the locomotion trajectories and learning uncertain rough terrain online. We construct both a footstep-by-footstep local navigation planner as well as a coarser global navigation planner which consider locomotion safety constraints.
    \item We develop the first ever planning framework that integrates Gaussian process models of unknown terrain elevation and motion perturbations for full-order bipedal locomotion. We propose a novel trajectory evaluation metric utilizing the GPs to minimize motion deviation and maximize information gain of the terrain estimation, thus increasing the feasibility of the navigation task. We benchmark the performance of multiple state-of-the-art GP kernels to evaluate their relative advantages for the bipedal navigation task.
    \item We evaluate the proposed methodology on simulations of a Digit bipedal robot in MuJoCo \cite{todorov2012mujoco}, demonstrating the validity of the reduced-order trajectories generated by our planner when implemented on the simulator using full-order robot dynamics.
\end{itemize}

\section{Preliminaries}\label{section: Preliminaries}

\subsection{Robot Model}\label{subsection: Robot Model}
We design our locomotion planner based on the Prismatic Inverted Pendulum Model (PIPM). PIPM has been proposed for agile, non-periodic locomotion over rough terrain~\cite{zhao2017robust} and integrated with Digit for navigation in partially observable environments and stair climbing tasks~\cite{shamsah2023integrated}.

Here we reiterate for completeness the mathematical formulation of our ROM. As shown in Figure \ref{fig:notation}, the CoM position $\boldsymbol{p}_{\rm com} = (x_{{\rm com}}, y_{{\rm com}}, z_{{\rm com}})^T$ is composed of the sagittal, lateral, and vertical positions in the global frame. We denote the apex CoM position as $\boldsymbol{p}_{{\rm apex}}=(x_{{\rm apex}},y_{{\rm apex}}, z_{{\rm apex}})^T$, the foot placement as $\boldsymbol{p}_{{\rm foot}}=(x_{{\rm foot}},y_{{\rm foot}}, z_{{\rm foot}})^T$, and $h_{\rm apex}$ is the relative apex CoM height with respect to the stance foot height. $v_{{\rm apex}}$ denotes the CoM velocity at $\boldsymbol{p}_{{\rm apex}}$. $\Delta y_1$ is the lateral distance between CoM and the high-level waypoint at apex. Formulating the dynamics for the next walking step as a hybrid control system
\begin{equation}\nonumber
    \Ddot{\boldsymbol{p}}_{{\rm com},n} = \begin{pmatrix}
\omega^2_{n}(x_{{\rm com}}-x_{{\rm foot},n})\\
\omega^2_{n}(y_{{\rm com}}-y_{{\rm foot},n})\\
a_n\omega^2_{n}(x_{{\rm com}}-x_{{\rm foot},n}) + b_n\omega^2_{n}(y_{{\rm com}}-y_{{\rm foot},n})
\end{pmatrix}
\label{eq:centrodial_dynamics}
\end{equation}
where the asymptote slope $\omega_n = \sqrt{g/z_{{\rm apex}, n}}$, and $z_{{\rm apex}, n} = a_n x_{{\rm foot},n} +b_n y_{{\rm foot},n} + h_{\rm apex} $. The hybrid control input is $\boldsymbol{u}_n=(\omega_{n},\boldsymbol{p}_{{\rm foot},n})$, with $\boldsymbol{p}_{{\rm foot},n}$ being a discontinuous input which creates a reset map.
\subsection{Phase-space Planning}
In phase-space planning (PSP), the sagittal CoM planning takes precedence over the lateral CoM planning. The decisions for the planning algorithm are primarily made in the sagittal phase-space, such as step length and CoM apex velocity, where we propagate the dynamics forward from the current apex state and backward from the next apex state until the two phase-space trajectories intersect. The intersection state defines the foot stance switching instant. On the other hand, the lateral phase-space parameters are searched for to adhere to the sagittal phase-space plan and have consistent timings between the sagittal and lateral plans. In this paper, we use the PSP method detailed in our previous work~\cite{shamsah2023integrated}.

\subsection{Gaussian Processes}\label{subsection: GP Definition}
In order to learn the uncertainties present in our bipedal system, we use Gaussian process (GP) regression:
\begin{definition}[Gaussian Process Regression]
\label{def:GP}%
Gaussian Process (GP) regression models a function $g_i:\mathbb{R}^n\to \mathbb{R}$ as a distribution with covariance $\kappa:\mathbb{R}^n\times\mathbb{R}^n\xrightarrow{}\mathbb{R}_{>0}$. Assume a dataset of $m$ samples $D = \{(\bm{\xi}^j,y_i^j)\}_{j\in\{1,...,m\}}$, where $\bm{\xi}^j\in\mathbb{R}^n$ is the input and $y^j_i$ is an observation of $g_i(\bm{\xi}^j)$ under Gaussian noise with variance $\sigma_{\nu_i}^2$. 
Let $K\in \mathbb{R}^{m\times m}$ be a kernel matrix defined elementwise by $K_{j\ell}=\kappa(\bm{\xi}^j,\bm{\xi}^\ell)$ and for $\bm{\xi}\in\mathbb{R}^n$, let $k(\bm{\xi})=[\kappa(\bm{\xi},\bm{\xi}^1) \;  \kappa(\bm{\xi},\bm{\xi}^2) \ldots $ $\kappa(\bm{\xi},\bm{\xi}^m)]^T\in \mathbb{R}^m$.
Then, the predictive distribution of $g_i$ at a test point $\bm{\xi}$ is the conditional distribution of $g_i$ given $D$, which is Gaussian with mean $\mu_{g_i,D}$ and variance $\sigma_{g_i,D}^2$ given by
\begin{align}
    \nonumber\mu_{g_i,D}(\bm{\xi}) &= k(\bm{\xi})^T(K+\sigma_{\nu_i}^2I_m)^{-1}Y\\
    \nonumber\label{Std Deviation} \sigma_{g_i,D}^2(\bm{\xi})&=\kappa(\bm{\xi},\bm{\xi})-k(\bm{\xi})^T(K+\sigma_{\nu_i}^2I_m)^{-1}k(\bm{\xi}),
\end{align}
where $I_m$ is the identity and $Y=\begin{bmatrix}y^1_i& y^2_i & \ldots & y^m_i\end{bmatrix}^T$.
\end{definition}
In practice, we use a sparse Gaussian process regression approximation \cite{leibfried2021tutorial} to reduce computational complexity. 

In this work, three different kernels for GP are benchmarked for predicting the terrain elevations, namely a radial basis function (RBF) kernel, a Neural Network (NN) kernel \cite{vasudevan2009gaussian}, and an Attentive Kernel \cite{chen2022ak}.

\subsubsection{RBF kernel}
The RBF kernel is a stationary kernel, commonly adopted in GP regression, in which the predictions are prone to be smooth and have the same degree of variability. The RBF kernel is defined as
\begin{equation}
    \nonumber\kappa(\bm{\xi}^i,\bm{\xi}^j) = \sigma_{f}^2 \exp\left(-\frac{\|\bm{\xi}^i - \bm{\xi}^j\|^2}{2\ell^2}\right),
\end{equation}
where \( \sigma_{f}^2 \) is signal variance and \( \ell \) is a lengthscale. 

\subsubsection{NN kernel} The NN kernel is nonstationary and resembles a neural network featuring a single hidden layer with infinitely many depth nodes and a sigmoid activation function \cite{neal2012bayesian}. 
This kernel models spatial correlation between data points which depends on the distance from the data origin until reaching a saturation region. The datapoints within the saturation region also correlate among themselves, allowing this kernel to adapt to rapid variations in the output values.

The work \cite{vasudevan2009gaussian} leverages the NN kernel for large-scale terrain elevation reconstruction, utilizing a local approximation method applied to the kernel. The approach involves implementing a KD-Tree to efficiently search for a specific number of nearest neighbors at each query point, facilitating the prediction of terrain elevation. This work demonstrates the suitability of the NN kernel for predicting terrain characterized by high variability in elevation and widely dispersed training data. This is due to the kernel's treatment of each query point as a data origin, correlating the output prediction solely with the training data nearest to that origin before reaching the saturation region. Within the saturation region, the kernel independently correlates other data points.

\begin{figure}[t]
    \centering
    \includegraphics[width=1.03\linewidth]{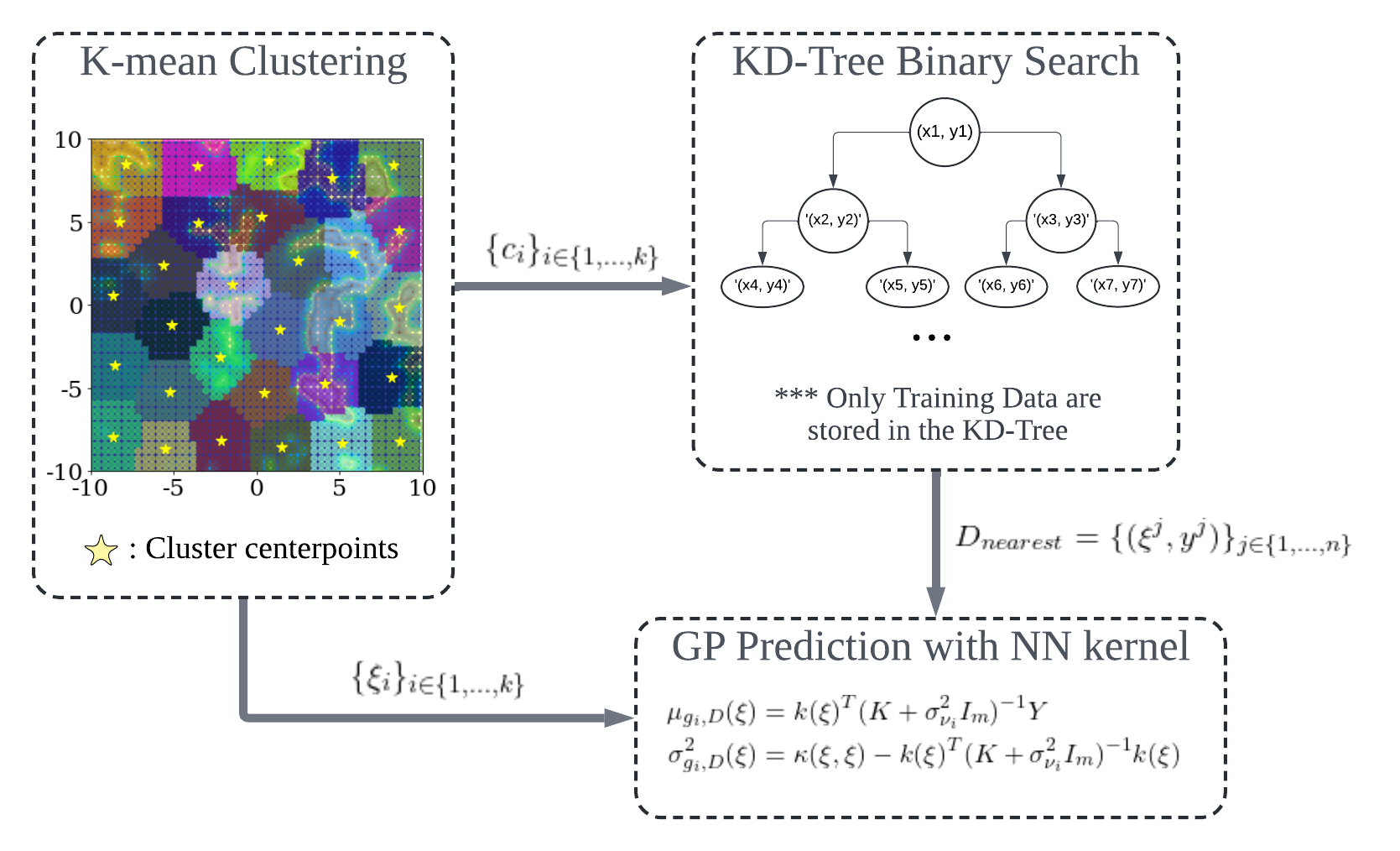}
    \caption{Overall block diagram of the local approximation method for GP prediction with NN kernel. Whenever the global elevation map undergoes re-estimation, all input locations in the terrain are split into \textit{k} clusters via K-mean clustering algorithm. Each cluster consists of a centerpoint $c_i$ and the subset input locations $\{\bm{\xi}\}_i$. Iterating through each of the \textit{k} clusters, the centerpoint is passed to the KD-Tree method to find \textit{n} nearest neighboring training datapoints denoted as $D_{nearest}^{i} = \{(\bm{\xi}^j,y^j)\}^{i}_{j\in\{1,...,n\}}$. Given \textit{n} closest training datapoints to the centerpoint (i.e., $D_{nearest}^{i}$) and the set of test points in that cluster to be estimated (i.e., $\{\bm{\xi}\}_i$), we then have all of the information to make GP predictions of the terrain heights at all input locations within the $i^{th}$ cluster region. With this local approximation method, GP prediction with NN kernel is only computed \textit{k} times, thereby reducing the computational load as compared to making a separate prediction for all input locations.}
    \label{fig:local_approx_block_diagram}
    \vspace{-0.2in}
\end{figure}

Despite the efficacy demonstrated by the terrain mapping method introduced in \cite{vasudevan2009gaussian}, incorporating such a local approximation technique into our framework entails significant computational overhead. The issue arises from the need to identify unique set of nearest neighbors for each query point across the terrain and individually compute predictions for each query point, thus significantly increasing the computation time. Consequently, we propose a  K-means clustering strategy to group closer input locations together and separate them into \textit{k} clusters. This approach ensures that input locations within the same cluster share identical sets of nearest neighboring training data, thereby mitigating the computational load. Detailed block diagram of the local approximation method is shown in Fig \ref{fig:local_approx_block_diagram}. In this study, we opt to partition our input space, comprising 2,500 locations (as detailed in \ref{section: Results}), into $k=200$ clusters to expedite terrain elevation prediction.

The NN kernel is defined as
\begin{align}\nonumber
    &k(\bm{\xi}^i,\bm{\xi}^j) = \\
    \nonumber&\sigma_{f}^2 \arcsin \bigg[\frac{ \beta + 2\bm{\xi}^{i^T} \Sigma \bm{\xi}^j}{\sqrt{(1 + \beta + 2\bm{\xi}^{i^T} \Sigma \bm{\xi}^i)(1 + \beta + 2\bm{\xi}^{j^T} \Sigma \bm{\xi}^j)}} \bigg]
\end{align}
where \(\Sigma = \begin{bmatrix} \ell_{x} & 0 \\ 0 & \ell{y} \end{bmatrix}^{-2} \), \( \beta\) is a bias factor, and \(\ell_{x} \text{ and } \ell_{y}\) are the lengthscales for input $x$ and $y$, respectively.

\subsubsection{Attentive Kernel} The Attentive Kernel is nonstationary and adaptive to different variability in the output by applying a weighted sum of many base RBF kernels with different lengthscales. The kernel also utilizes the Instance Selection method, whereby each input location is assigned a membership vector to ensure that the training input within the same neighborhood can break their correlation given abrupt changes in the training output \cite{chen2022ak}. The incorporation of weighted summation and instance selection allows the Attentive kernel to achieve robust predictive performance in the face of terrain elevation variability, while also enhancing its capability for uncertainty quantification. Both weights and membership vectors are learned from the training dataset using a two-hidden-layer neural network with TanH activation function. 
The kernel is defined as 

\begin{equation}
    \nonumber\kappa(\bm{\xi}^i,\bm{\xi}^j) = \alpha \bar{z}^T \bar{z}' \sum_{m=1}^{M} \bar{w}_m \kappa_m(\bm{\xi}^i, \bm{\xi}^j) \bar{w}_m'
\end{equation}
where \(\alpha\) is the amplitude constant, \( \bar{w} \) and \( \bar{z} \) are the weight and membership vector respectively, and \( \{\kappa(\bm{\xi}^i,\bm{\xi}^j)\}_{m=1}^M \) are base RBF kernels with varying lengthscales.

\section{Problem Statement}\label{Section: Problem}
We now formally define the problem we study in this work. Consider an environment in which the terrain elevation is uncertain, creating multiple challenges for bipedal locomotion. First, regions with high terrain elevation may be untraversable, creating obstacles in the environment. Additionally, the terrain elevation is an input to the PSP model, so inaccurate terrain estimations increase deviation and create instability in planned footstep trajectories. Thus, we incorporate an additional objective in the navigation planner to learn the terrain online, improving the dynamical feasibility of planned trajectories.

The primary objective is for the robot to reach a desired location in the environment. However, the terrain elevation is initially unknown, creating untraversable regions and perturbing the motion of the robot. Thus, the robot must sample the environment to learn an accurate representation of the terrain map and feasibly traverse towards the goal. There also exists motion perturbations resulting from the model error between the PIPM used for planning and the full-order dynamics, which complexifies the overall uncertainty learning problem. 

\noindent\textbf{Problem Statement:} Design a hierarchical planning framework for a bipedal robot which generates dynamically feasible trajectories to reach a desired goal location in an environment with unknown terrain features. Learn online the terrain elevation and the resulting motion perturbations in order to avoid untraversable regions and minimize the error between the desired motion plans and the measured trajectories from a full-body robot dynamic simulation.
\begin{figure}[t]
    \centering
    \includegraphics[width=0.85\linewidth]{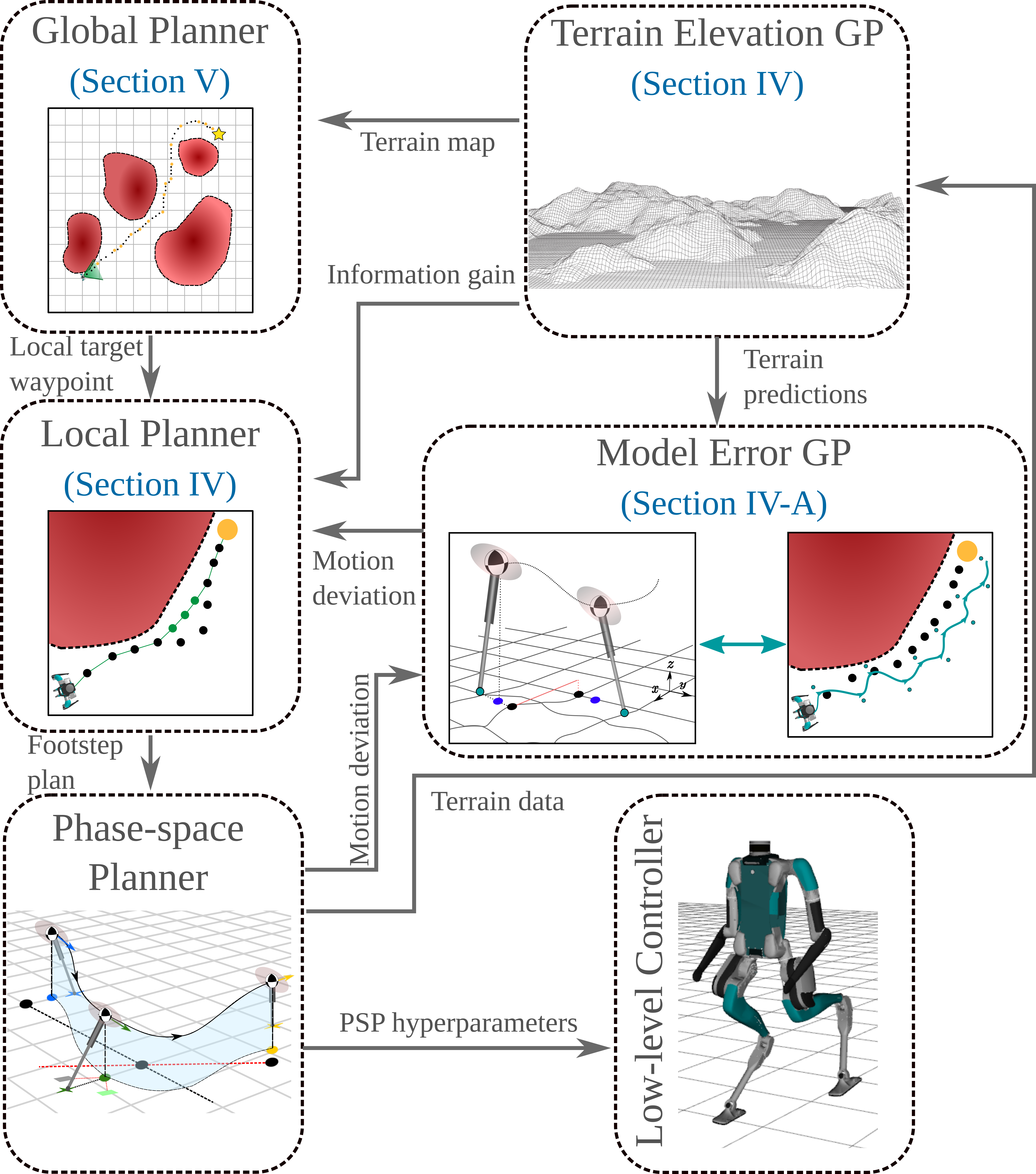}
    \caption{Overall block diagram of the proposed global-local planning framework for bipedal navigation over rough terrain.}
    \label{fig:block_diagram}
    \vspace{-0.2in}
\end{figure}

Our approach to this problem is as follows. We first learn the unknown terrain elevation and characterize its effect on the motion of the robot using GPs. Then, we design a local navigation planner which generates dynamically feasible waypoints to avoid untraversable regions and balance objectives of environmental exploration and motion perturbation minimization. Subsequently, we develop a global navigation planner which generates near-horizon targets for the local navigation planner which guide the robot towards the global goal while ensuring dynamically feasible plans. Figure \ref{fig:block_diagram} depicts the structure of the proposed approach.
\section{Local Navigation Planner}\label{section: Local Planner}
In this section, we propose a local navigation planner to generate a footstep-by-footstep motion plan.
We first define local navigation trajectories:
\begin{definition} [Local Navigation Trajectory]
   A local navigation trajectory $\mathcal{A}_{\boldsymbol{w} \to \boldsymbol{w}'}$ from a start waypoint $\boldsymbol{w}=(x,y,\theta)$ to an end waypoint $\boldsymbol{w}'$ is an $n$-step sequence $\{a_{HL,0},\cdots,a_{HL,n-1}\}$ of high level actions $a_{HL,i}=(d_i,\Delta\theta_i,\Delta z_i,\psi_i)$. The sequence $\mathcal{A}_{\boldsymbol{w} \to \boldsymbol{w}'}$ induces a set of apex CoM waypoints $\mathcal{W}(\mathcal{A}_{\boldsymbol{w} \to \boldsymbol{w}'})=\{\boldsymbol{w}_0,\cdots,\boldsymbol{w}_{n-1}\}$ such that $\boldsymbol{w}_0=\boldsymbol{w}$, $\boldsymbol{w}_{n-1}=\boldsymbol{w}'$, and $\boldsymbol{w}_{i+1}=\boldsymbol{w}_i+\begin{bmatrix}
       d_i\cos(\sum_{j=0}^i\Delta\theta_j+\theta_0),d_i\sin(\sum_{j=0}^i\Delta\theta_j+\theta_0),\Delta\theta_i
   \end{bmatrix},$ $ \forall i \in \{0,\cdots,{n-2}\}$.
\end{definition}
The local navigation trajectory parameters are illustrated in Figure \ref{fig:local_planner}(c).
Given a target endpoint generated by a global navigation planner (detailed in Section \ref{section: Global Planner}), the local navigation planner generates a plan which seeks to reach the target while balancing two objectives. First, the plan encourages exploration so that the robot can more accurately characterize terrain elevation, improving the feasibility of future motion plans. Second, the plan designs a sequence of waypoints which minimize the expected motion perturbation to reach the desired target.

\subsection{Gaussian Process Learning of Terrain and Model Errors}\label{subsection: GP Methodology}
We first detail the GP structure we use to learn the unknown terrain elevation and the motion perturbations resulting from both terrain and model errors. This structure builds on our previous GP modeling work in \cite{jiang2023abstraction}.

We use a terrain GP $\hat{z}(x,y)$ for which the input is a global location $(x,y)$ and the output is the corresponding terrain height at that location, $z$. The kernel of the GP is either the RBF, NN, or Attentive Kernel discussed in Section \ref{subsection: GP Definition}. We evaluate the respective performance of these approaches in Section \ref{section: Results}. At runtime, the GP is trained and updated using data collected as the robot traverses through the environment. The mean prediction of the terrain GP is used in the PSP controller to generate feasible footstep trajectories, and the variance is utilized to determine the information gain along each local trajectory, as discussed later in Section \ref{subsection: Locomotion-aware Local RRT*}.

Given the GP model of terrain elevation, we characterize the model error in the lateral direction at each step using the GP $\Delta \hat{y}_{1}(d_{c},\Delta\theta_{c},\Delta z_{c},d_n,\Delta\theta_n,\Delta z_n)$. The input parameters $d_c,\Delta\theta_c,\Delta z_c$ represent the difference in distance, heading angle, and terrain elevation, respectively, between the previous waypoint and the current waypoint at a given step. The parameters with subscript $n$ represent the same values measured between the current waypoint and the next waypoint. The PSP controller is designed to achieve the desired sagittal distance exactly and then minimize lateral error as a secondary objective. Thus, we choose to focus on modeling lateral deviation. 

The model error GP is trained offline on a dataset generated by simulating steps over the entire range of input parameters using the low-level controller in Section \ref{section: Results} and measuring the resulting motion perturbation. In practice, the stance of the robot also affects the motion perturbation. Steps taken with the left foot result in lateral deviation to the left of the waypoint (in the local robot frame), whereas steps taken with the right foot deviate to the right of the targeted waypoint. For efficient learning, we learn the absolute value of the lateral deviation with a single model error GP taking into account both left and right footsteps. Then, when calling the model error GP, we assign a positive value for the deviation on left footsteps, and we assign a negative value for the deviation on right footsteps.

The local navigation planner requires predictions of the expected model error for proposed waypoint sequences. To obtain these predictions, we call the model error GP $\Delta \hat{y}_{1}$ for each step of the sequence. The GP input parameters $d,\Delta\theta$ can be extracted directly from the waypoint sequence, but the parameter $\Delta z$ depends on the unknown terrain elevation at each waypoint. Thus, we choose to use the mean $\mu_{\hat{z}}$ of the terrain elevation GP prediction at the waypoints to approximate $\Delta z$ at each step. Since each output $\Delta \hat{y}_{1}$ is in the local frame w.r.t. a specific  waypoint, we apply coordinate transforms on the outputs to place them in the global frame.

\subsection{Locomotion-dynamics-aware Local RRT*}\label{subsection: Locomotion-aware Local RRT*}
\begin{figure}[t]
    \centering
    \includegraphics[width=0.85\linewidth]{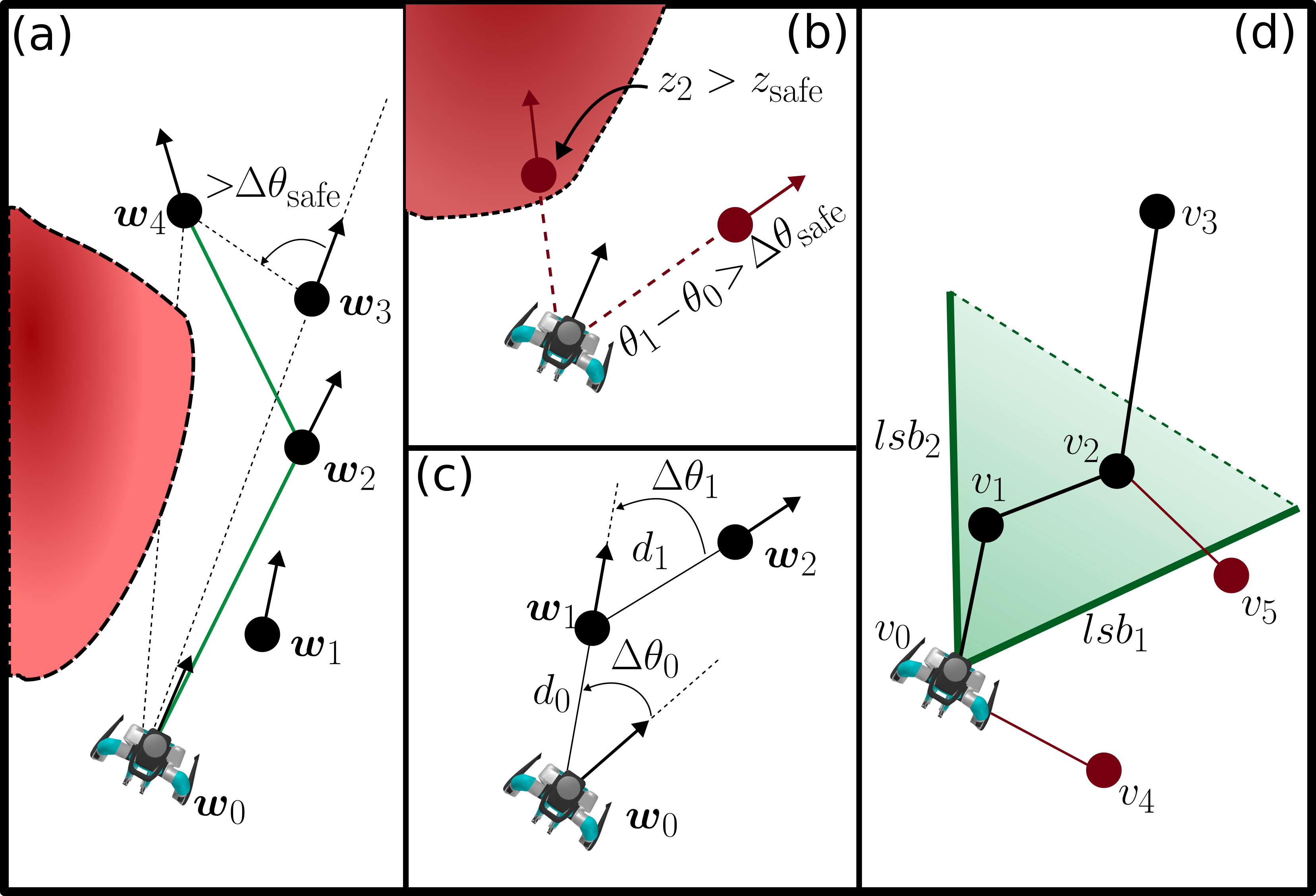}
    \caption{(a) Illustration of the smoothing algorithm. The goal is to find the smoothest path between the start $\boldsymbol{w}_0$ and the target $\boldsymbol{w}_4$. Connecting $\boldsymbol{w}_0$ and $\boldsymbol{w}_4$ directly is invalid because the line between them passes through an obstacle. Connecting to $\boldsymbol{w}_3$ is also invalid because the resultant heading angle exceeds the limit $\Delta\theta_{{\rm safe}}.$ Thus, the smoothing algorithm connects $\boldsymbol{w}_0$ to $\boldsymbol{w}_2$ to $\boldsymbol{w}_4$. (b) Illustration of LDA-L-RRT* vertex selection criteria. The left red vertex is invalid because it lies in an obstacle, and the right red vertex is impossible because the heading angle change exceeds $\Delta\theta_{{\rm safe}}$. (c) Illustration of the local navigation trajectory parameters. (d) Illustration of the LDA-G-RRT* safety criteria. The solid green lines are the locomotion safety barriers, and the shaded green area is their convex hull. Starting from vertex $v_0$, the connection to $v_4$ is invalid because $v_4$ is outside the locomotion safety barriers. The connection from $v_2$ to $v_4$ is invalid as it crosses a locomotion safety barrier. The connection from $v_2$ to $v_3$ is valid, as the dashed green line is not a locomotion safety barrier.}
    \label{fig:local_planner}
\end{figure}

We now define the locomotion-dynamics-aware RRT* algorithm we use for local planning:
\begin{definition}[Locomotion-dynamics-aware Local RRT*]
    The LDA-L-RRT* algorithm modifies the standard RRT* algorithm by placing additional constraints on new vertices in the search as follows. First, the configuration of a vertex is $\boldsymbol{w}=(x_{{\rm apex}},y_{{\rm apex}},\theta)$, where $(x_{{\rm apex}},y_{{\rm apex}})$ is the planar apex position and $\theta$ is the heading angle. Then, consider one step of the standard RRT* algorithm in which a random point in the environment $(x_{{\rm rand}},y_{{\rm rand}})$ is selected, for which the nearest vertex is $\boldsymbol{w}_1=(x_{\rm apex,1},y_{\rm apex,1},\theta_1)$. A candidate vertex $\boldsymbol{w}'=(x_{{\rm apex}}',y_{{\rm apex}}',\theta')$ is calculated as
    \begin{align}\nonumber
        &\begin{bmatrix}x_{{\rm apex}}' & y_{{\rm apex}}'\end{bmatrix} = \\
        \nonumber &d_{{\rm safe}}\frac{\begin{bmatrix}x_{{\rm rand}} & y_{{\rm rand}}\end{bmatrix}-\begin{bmatrix}x_{\rm apex,1} & y_{\rm apex,1}\end{bmatrix}}{\left\|\begin{bmatrix}x_{{\rm rand}} & y_{{\rm rand}}\end{bmatrix}-\begin{bmatrix}x_{\rm apex,1} & y_{\rm apex,1}\end{bmatrix}\right\|_2}, \\
        \nonumber &\theta' = \arctan\bigg(\frac{y_{{\rm rand}}-y_{\rm apex,1}}{x_{{\rm rand}}-x_{\rm apex,1}}\bigg),
    \end{align}
    where $d_{\rm safe}$ is a safe step distance determined as in \cite[Theorem IV.1]{shamsah2023integrated}. Then, the candidate $\boldsymbol{w}'$ is added to the graph if and only if it satisfies the following conditions:
    \begin{enumerate}
        \item The heading angle change between connected vertices is less than a dynamically feasible limit $\Delta\theta_{\rm safe}$ calculated as in \cite[Theorem IV.2]{shamsah2023integrated}:
        \begin{equation}
            \nonumber|\theta'-\theta_1|\leq\Delta\theta_{{\rm safe}},
        \end{equation}
        \item The GP predicted terrain elevation at $\boldsymbol{w}'$ is smaller than a dynamically feasible limit $z_{{\rm safe}}$:
        \begin{equation*}
            \mu_{\hat{z}}(x_{{\rm apex}}',y_{{\rm apex}}')\leq z_{{\rm safe}}.
        \end{equation*}
    \end{enumerate}
    
    If $\boldsymbol{w}'$ does not satisfy the conditions above, a new candidate $\boldsymbol{w}'$ is calculated by performing the above procedure with the same random point $(x_{{\rm rand}},y_{{\rm rand}})$ and the next closest vertex in the graph $\boldsymbol{w}_2$. The process continues through all of the vertices in the current graph until a candidate vertex is successfully added to the graph. If there is no node that satisfies both conditions, the nearest vertex $\boldsymbol{w}_{{\rm near}}$ to $(x_{{\rm rand}},y_{{\rm rand}})$ that does not have a child node is identified. A final candidate vertex $\boldsymbol{w}''$ is proposed with values
    \begin{align}
        \nonumber \begin{bmatrix}x_{{\rm apex}}'' \\ y_{{\rm apex}}''\end{bmatrix} &= \begin{bmatrix}x_{{\rm apex, near}} \\ y_{{\rm apex, near}}\end{bmatrix} + d_{{\rm safe}}\begin{bmatrix}\cos(\theta_{{\rm near}} + \Delta\theta_{{\rm safe}})\\ \sin(\theta_{{\rm near}} + \Delta\theta_{{\rm safe}})\end{bmatrix}, \\
        \nonumber\theta'' &= \theta_{{\rm near}}+\Delta\theta_{{\rm safe}}.
    \end{align}
    This candidate vertex always satisfies condition 1), so it is added to the graph if it also satisfies condition 2). If not, then no vertex is added to the graph in the current step.
\end{definition}

With these conditions, the waypoint sequences generated by the LDA-L-RRT* algorithm are guaranteed to be dynamically feasible with respect to the PIPM safety conditions proposed in \cite{shamsah2023integrated} and will avoid untraversable regions with excessive terrain elevation. However, the resulting trajectories can change heading angle rapidly at each step, directly increasing motion errors. Thus, we propose a trajectory-smoothing algorithm to smooth the LDA-L-RRT* trajectories to improve trajectory tracking performance. The idea of the smoothing algorithm is to replace the ``zigzag"-prone trajectories typical of RRT-generated trajectories with straight lines more amenable to bipedal locomotion. The algorithm begins at the starting waypoint $\boldsymbol{w}_0 = (x_{\rm apex,0},y_{\rm apex,0},\theta_0)$ of a LDA-L-RRT* motion plan and finds the furthest waypoint $\boldsymbol{w}_i = (x_{\rm apex,i},y_{\rm apex,i},\theta_i)$ along the trajectory which satisfies the following conditions:
\begin{enumerate}
    \item There is no untraversable terrain along the line connecting $\boldsymbol{w}_0$ and $\boldsymbol{w}_i$:
    \begin{align}
        \nonumber&\mu_{\hat{z}}(x_{{\rm apex}}',y_{{\rm apex}}')\leq z_{{\rm safe}}\\\
        \nonumber&\forall (x_{{\rm apex}}',y_{{\rm apex}}')\in conv(\boldsymbol{w}_0,\boldsymbol{w}_i),
    \end{align}
    where $conv()$ is the convex hull, i.e., minimal convex set containing the two points.
    \item The heading angle change between $\boldsymbol{w}_0$ and $\boldsymbol{w}_i$ and between $\boldsymbol{w}_i$ and $\boldsymbol{w}_{i+1}$ is valid:
    \begin{equation}
        \nonumber|\theta_0-\theta_i|\leq\Delta\theta_{{\rm safe}},|\theta_i-\theta_{i+1}|\leq\Delta\theta_{{\rm safe}}
    \end{equation}
\end{enumerate}

Once $\boldsymbol{w}_i$ is identified, a new sequence of waypoints with appropriate step lengths is generated on the line between $\boldsymbol{w}_0$ and $\boldsymbol{w}_i$, replacing the waypoints $\{\boldsymbol{w}_1,\cdots,\boldsymbol{w}_{i-1}\}$. Then, the smoothing algorithm continues from $\boldsymbol{w}_i$, repeating until the target waypoint $\boldsymbol{w}_\ell$ is reached.
\begin{figure}[t]
    \centering
    \includegraphics[width=0.95\linewidth]{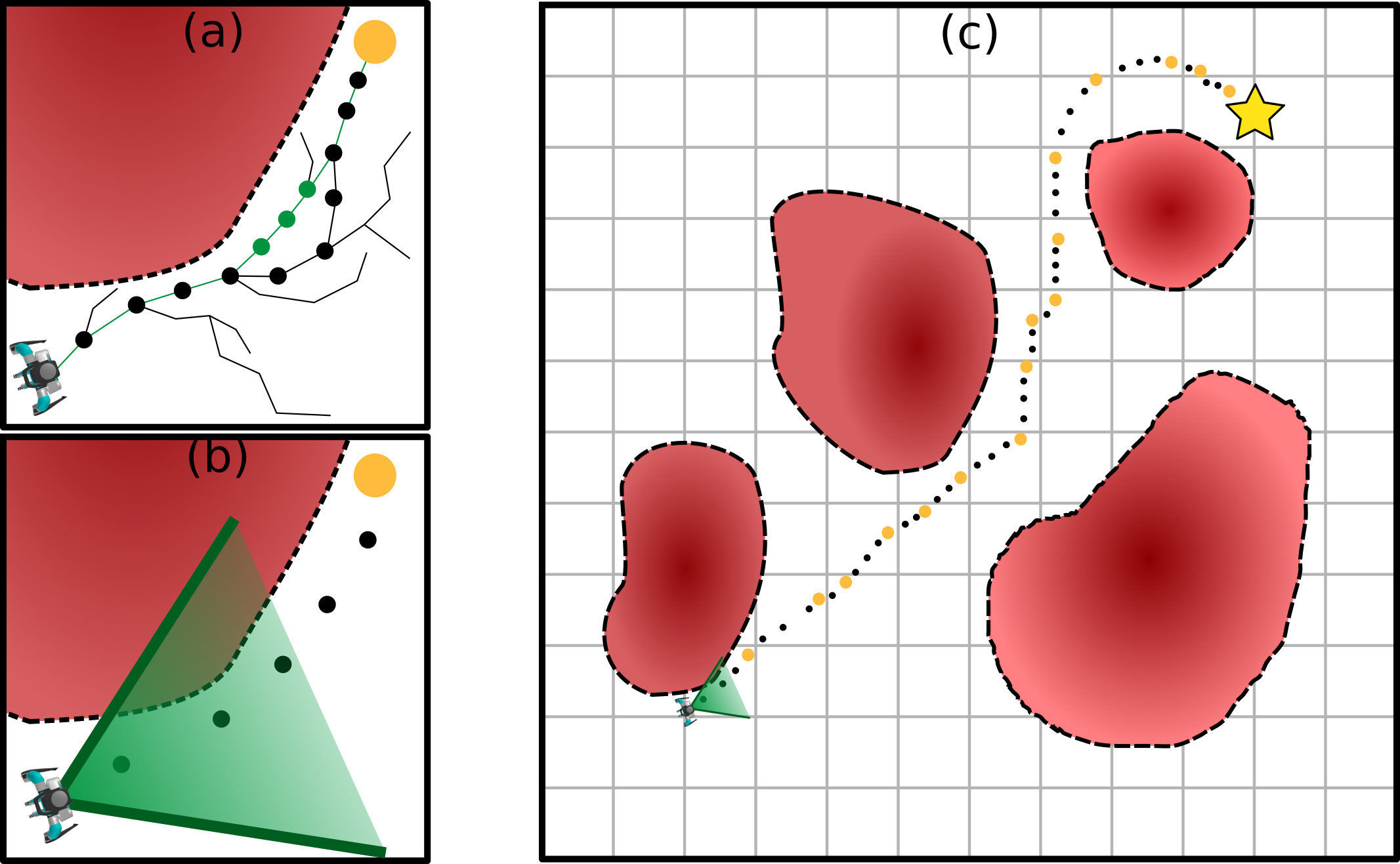}
    \caption{(a) Trajectory generated by the LDA-L-RRT* algorithm for reaching a local waypoint. The green dots show the trajectory modifications made by the proposed smoothing algorithm. (b) Illustration of the locomotion safety barrier region around the start vertex in the global navigation planner (a zoomed-in version of the one shown in (c)). (c) Trajectory generated by the LDA-G-RRT* algorithm. The black and orange dots combined represent the planned trajectory, from which the orange dots are selected as local waypoints. The green triangle indicates the locomotion safety barrier region constraining vertices near the start point.}
    \label{fig:global_rewiring}
    \vspace{-0.2in}
\end{figure}
Figure \ref{fig:local_planner}(a) illustrates the smoothing algorithm, Figure \ref{fig:local_planner}(b) shows the vertex safety constraints, and Figure \ref{fig:global_rewiring}(a) shows a conceptual example of the waypoint sequence generated by LDA-L-RRT*.

For each local target, we run the LDA-L-RRT* algorithm $m$ times to generate candidate trajectories $\{\mathcal{A}_{\boldsymbol{w} \to \boldsymbol{w}',j}\},j\in\{1,\cdots,m\}$. We then select an optimal trajectory $\mathcal{A}_{\boldsymbol{w} \to \boldsymbol{w}'}^*$ using the formula
\begin{align}
    &j^* = \nonumber\argmax_j [ -\alpha({\rm error}(\mathcal{A}_{\boldsymbol{w} \to \boldsymbol{w}',j}))+\beta ({\rm info}(\mathcal{A}_{\boldsymbol{w} \to \boldsymbol{w}',j}))], \\
    \nonumber&{\rm error}(\mathcal{A}_{\boldsymbol{w} \to \boldsymbol{w}',j}) = \sum_{ a_{HL} \in \mathcal{A}_{\boldsymbol{w} \to \boldsymbol{w}',j}}T(\hat{y}_{1}(d,\Delta\theta, \Delta z)) \\
    \nonumber&{\rm info}(\mathcal{A}_{\boldsymbol{w} \to \boldsymbol{w}',j}) = \sum_{\boldsymbol{w}_i \in \mathcal{W}(\mathcal{A}_{\boldsymbol{w} \to \boldsymbol{w}',j})} \frac{1}{2} log(2\pi\sigma^2_{\hat{z}}(\boldsymbol{w}_i)) + \frac{1}{2},
\end{align}
where $\alpha,\beta\in\mathbb{Z}_{\geq 0}$ and $T$ is the transform from a local waypoint frame to the global frame. The optimal solution is $\mathcal{A}_{\boldsymbol{w} \to \boldsymbol{w}'}^* = \mathcal{A}_{\boldsymbol{w} \to \boldsymbol{w}',j^*}$. Intuitively, the optimal path minimizes the $error$ predicted by the model error GP over the waypoint sequence and maximizes the information gain $info$, which rewards the traversal of areas currently having high uncertainty in the terrain elevation GP. The parameters $\alpha,\beta$ tune the importance of these two objectives.
 
\section{Global Navigation Planner}\label{section: Global Planner}
In a large, global environment, it is computationally expensive to perform footstep-by-footstep local planning in order to reach a global goal. Thus, in this section we propose a coarse global navigation planner which plans waypoints as inputs for the LDA-L-RRT* algorithm as described in Section \ref{section: Local Planner}. This global navigation planner guides the robot towards the global goal. Then, we detail the complete local-global planning framework.
\subsection{Locomotion-dynamics-aware Global RRT*}
For the global navigation planner, we propose another modified RRT* algorithm which incorporates bipedal locomotion constraints while reducing computational costs as compared to the local navigation planner of Section \ref{section: Local Planner} for planning in large environments.
\begin{definition}[Locomotion-dynamics-aware Global RRT*]
    The locomotion-dynamics-aware global RRT* (LDA-G-RRT*) algorithm modifies the standard RRT* algorithm by placing additional constraints on new vertices in the search as follows. First, we partition the global environment into hyper-rectangular regions $\{W_q\}_{q\in Q}$:
    \begin{equation}
    \label{eq:part}
        \nonumber W_q=\{(x_{{\rm com}},y_{{\rm com}})\mid \ \underline{x}_{q}\leq x_{{\rm com}}\leq \overline{x}_{q},\underline{y}_{q}\leq y_{{\rm com}}\leq \overline{y}_{q}\},
    \end{equation}
    where the inequality is taken elementwise for lower and upper bounds $\underline{\cdot}_{q},\overline{\cdot}_{q}\in\mathbb{R}$ and $Q$ is a finite index set of the regions. The configuration of a vertex is $v=(x_{{\rm com}},y_{{\rm com}})$. Additionally, for the starting vertex $v_0=(x_0,y_0)$ we know the heading angle $\theta_0$ from the robot's current state. We create locomotion safety barriers around the start vertex, defined as
    \begin{align}\nonumber
        lsb_1 &= \bigg\{\begin{bmatrix}
            x_0 \\
            y_0
        \end{bmatrix}+2d_{{\rm step}}\gamma\begin{bmatrix}
            \cos(\theta_0-\Delta\theta_{{\rm safe}}) \\
            \sin(\theta_0-\Delta\theta_{{\rm safe}})
        \end{bmatrix}\bigg\},\gamma\in[0,1], \\
        \nonumber
        lsb_2 &= \bigg\{\begin{bmatrix}
            x_0 \\
            y_0
        \end{bmatrix}+2d_{{\rm step}}\gamma\begin{bmatrix}
            \cos(\theta_0+\Delta\theta_{{\rm safe}}) \\
            \sin(\theta_0+\Delta\theta_{{\rm safe}})
        \end{bmatrix}\bigg\},\gamma\in[0,1],
    \end{align}
    where $d_{{\rm step}}$ is a desired distance between vertices.
    
    Then, consider one step of the standard RRT* algorithm in which a random point in the environment $(x_{{\rm rand}},y_{{\rm rand}})$ is selected, for which the nearest vertex is $v$. A candidate vertex $v'=(x',y')$ is calculated as
    \begin{align}\nonumber
        \begin{bmatrix}x' & y'\end{bmatrix} &= d_{{\rm step}}\frac{\begin{bmatrix}x_{{\rm rand}} & y_{{\rm rand}}\end{bmatrix}-\begin{bmatrix}x & y\end{bmatrix}}{\left\|\begin{bmatrix}x_{{\rm rand}} & y_{{\rm rand}}\end{bmatrix}-\begin{bmatrix}x & y\end{bmatrix}\right\|_2}
    \end{align} 
    Then, the candidate $v'$ is added to the graph if and only if it satisfies the following conditions:
    \begin{enumerate}
        \item Any vertex connected to the starting vertex $v_0$ must be in the convex hull $conv\{lsb_1,lsb_2\}$.
        \item Any connection between vertices in the graph must not intersect a locomotion safety barrier.
        \item The GP predicted terrain elevation at $v'$ is smaller than a dynamically feasible limit $z_{{\rm safe}}$:
        \begin{equation*}
            \mu_{\hat{z}}(x',y')\leq z_{{\rm safe}}.
        \end{equation*}
    \end{enumerate}
    Figure \ref{fig:local_planner}(d) illustrates the locomotion safety barriers.
    
    Once a sequence of vertices $\{v_0,v_1,\cdots,v_n\}$ from $v_0$ to the desired target position $v_n$ has been found, we generate a sequence of corresponding local waypoints recursively as follows. We start at $v_0$ and identify the largest index $i$ such that the vertices $\{v_0,v_1,\cdots,v_i\}$ are all elements of the same hyper-rectangular region $W_0$. The vertex $v_i$ is the first local waypoint. Then, we move to vertex $v_{i+1}$ and find the largest index $j$ such that the vertices $\{v_i,v_1,\cdots,v_j\}$ are all elements of the same hyper-rectangular region $W_i$, adding $v_j$ to the sequence of local waypoints. We repeat this process until the target $v_n$ has been added to the sequence of local waypoints and return the complete sequence.
\end{definition}

The purpose of the locomotion safety barriers is to ensure that the first several steps of the robot are feasible with respect to the heading angle change for each step.

In practice, we select the step size $d_{{\rm step}}$ for the LDA-G-RRT* algorithm larger than the step size $d_{{\rm safe}}$ for the local LDA-L-RRT* algorithm. Additionally, the locomotion safety barrier constraints on vertex selection for LDA-G-RRT* only hold within a small radius of the starting vertex $v_0$, whereas LDA-L-RRT* constrains every vertex. These two features result in the computational efficiency of LDA-G-RRT*.
\begin{algorithm}[t]
\caption{Global-Local Planning Framework}\label{alg:Local-Global Planning Framework}
\KwIn{Start waypoint $\boldsymbol{w}_0$, target waypoint $\boldsymbol{w}_t$}
\SetKw{Init}{Initialize}
\Init{Terrain GP $\hat{z}(x,y)$}\;
\Init{Model Error GP $\Delta\hat{y}_{1}$}\;
\Init{Current position $\boldsymbol{w}_c=\boldsymbol{w}_0$}\;
\While{$\boldsymbol{w}_c\neq \boldsymbol{w}_t$}{
    Run LDA-G-RRT* algorithm with target $x_t$ and obtain local target waypoint $\boldsymbol{w}_\ell$\;
    Run LDA-L-RRT* with target $\boldsymbol{w}_\ell$ and obtain footstep plan\;
    Execute footstep plan and collect terrain data $\{(x_i,y_i),z_i\}$ along the trajectory\;
    Update current waypoint $\boldsymbol{w}_c$\;
    Retrain terrain GP $\hat{z}$ on collected data\;
}
\end{algorithm}
Figure \ref{fig:global_rewiring}(b),(c) illustrates LDA-G-RRT*.
\subsection{Overall Framework}
We now detail the overall planning framework. We assume that there exists \textit{a priori} a small dataset of terrain elevation data points, and we initialize the terrain GP by training on this dataset. This assumption can be relaxed by, \textit{e.g.}, initializing the hyperparameters of the terrain GP to have a conservative level of uncertainty throughout the environment. The model error GP is trained offline as in Section \ref{subsection: GP Methodology}. We then run the LDA-G-RRT* algorithm to obtain a local target waypoint, which we send to the LDA-L-RRT* algorithm to generate a footstep-by-footstep motion plan. The robot executes this motion plan, collecting terrain elevation data along the trajectory. Once the robot reaches the local target waypoint, the terrain GP is retrained on the newly collected data. Then, LDA-G-RRT* is run again and the rest of the process repeats until the global target waypoint is reached. The complete framework is summarized in Algorithm \ref{alg:Local-Global Planning Framework}.
\begin{figure}[t]
    \centering
    \includegraphics[width=0.95\columnwidth]{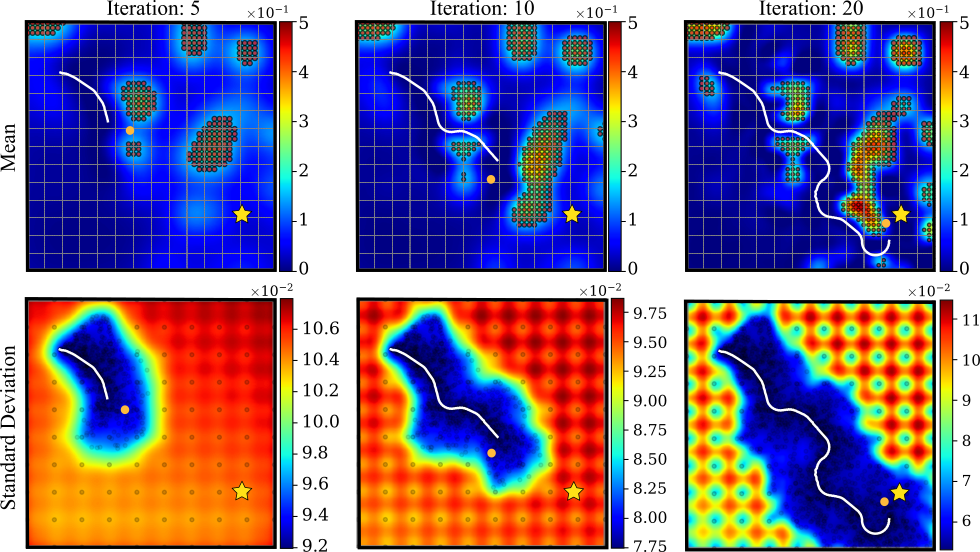}
    \caption{Snapshots of the state of the system at various points throughout one sample run. In all plots, the white line is the CoM trajectory, the orange dot is the local target waypoint, and the yellow star is the global goal. In the left plot, which is early in the trajectory, the robot has only characterized the terrain in a small region around the start point. In the middle plot, the robot has safely passed through the first set of obstacles based on the earlier terrain data gathered, and it has learned most of the large obstacle blocking the goal. Finally, the last plot shows the completed run, where the robot has learned the terrain sufficiently to reach the goal.}
    \label{fig:storyboard}
\end{figure}

\section{Results}\label{section: Results}
\begin{figure*}[t]
    \centering
    \subfigure[N37W112]{\includegraphics[width=0.3\textwidth]{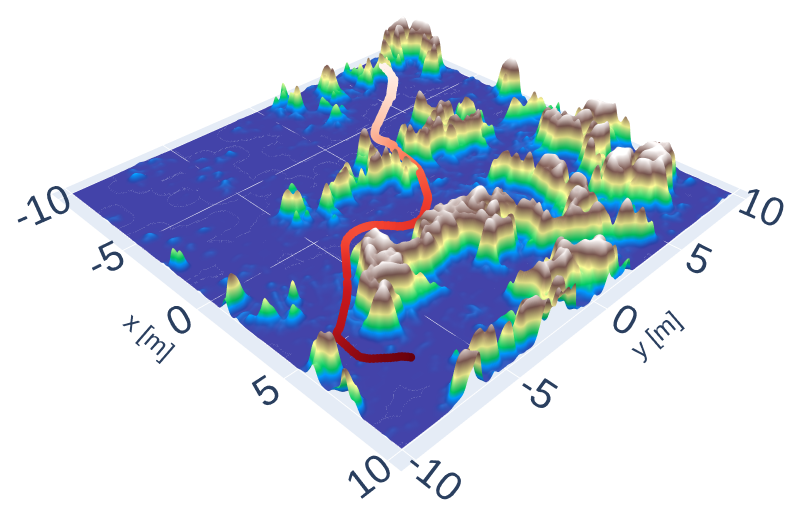}}
    \subfigure[N24W102]{\includegraphics[width=0.3\textwidth]{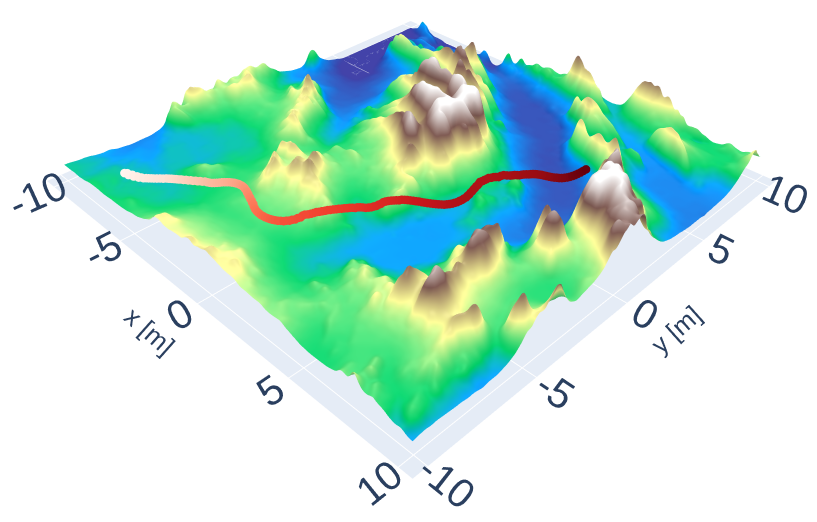}} 
    \subfigure[N17E102]{\includegraphics[width=0.3\textwidth]{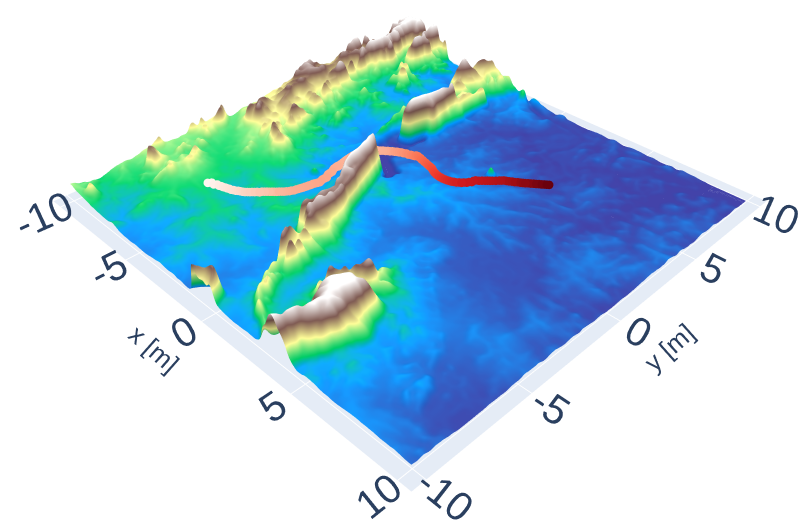}} 
    \caption{Three environments used in the navigational task case study with sample final trajectory (start from white to red).}
    \label{fig:terrain_visual}
\end{figure*}

We evaluate our framework on simulations of a Digit bipedal robot navigating three environments with varying terrain (i.e., N37W112, N24W102, N17E102) as shown in Figure \ref{fig:terrain_visual}. Each of the three terrains exhibits distinct levels of nonstationarity. The N37W112 terrain is characterized by numerous standalone hills that render it impassable for the bipedal robot, presenting a high degree of nonstationarity marked by many abrupt changes in elevation. In contrast, the N24W102 terrain features fewer significant changes in elevation, displaying a more gradual form of nonstationarity. Finally, the N17E102 terrain exhibits the least nonstationarity, consisting of only one elongated elevation across the landscape. Each environment is $20\times20$ meters in size, with terrain elevation varying between 0 and 0.5 meters.
\begin{figure*}[t]
    \centering
    \includegraphics[width=0.99\textwidth]{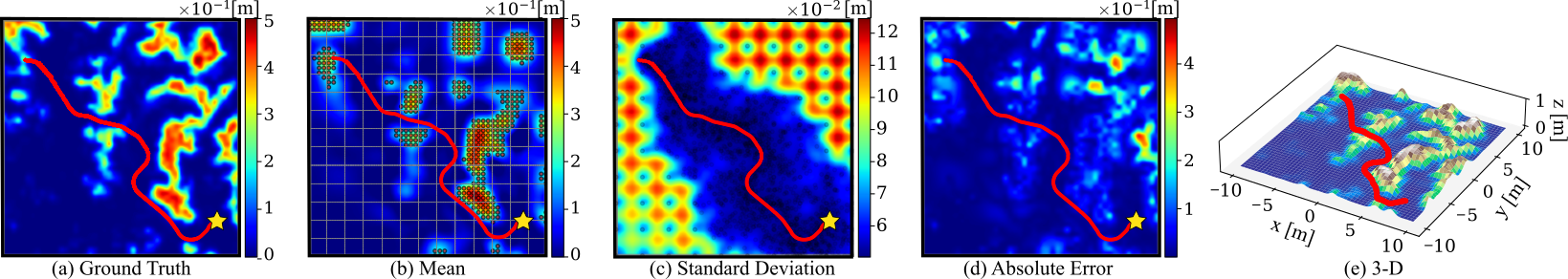}
    \caption{A sample run of our proposed framework using an Attentive-Kernel-based GP on the terrain N37W112. In each graph, the red line represents the CoM trajectory; the yellow dots are the discrete footsteps; and the yellow star is the global goal location. (a) The ground truth elevation map with the heatmap. Regions with elevation above 0.15 meters are untraversable. (b) The final GP estimation of the terrain for the Attentive Kernel method. (c) The standard deviation of the terrain GP. Higher values indicate a higher uncertainty. Note that, as expected, the robot has a high confidence around its actual trajectory and low confidence far from where it has sampled. (d) The absolute error of the terrain GP (b) compared to the ground truth (a). The estimation shows high accuracy near the actual trajectory, and the robot does not need an accurate estimation of terrain far away from its trajectory in order to reach the goal. (e) A 3D view of the Digit robot navigation trajectory through the environment.}
    \label{fig:results}
\end{figure*}

The terrain GP $\hat{z}$ model is initialized on prior information consisting of 500 randomly-selected points across the terrain (out of the total of 2500 points). Then, the GP model is updated online as new data is collected as the robot traverses the environment. The sensor is programmed to simulate the collection of 10 sample points within a 3-meter radius of the robot during each step. The model error GP $\Delta\hat{y}_{1}$ is trained offline as described in Section \ref{subsection: GP Methodology}. Across all simulations, the average motion perturbation per step was 1.75e-2 meters, and the average prediction error of $\Delta\hat{y}_1$ per step was 2.06e-4 meters. Thus, $\Delta\hat{y}_1$ accurately evaluates the motion perturbation. Finally, for the LDA-L-RRT* local navigation planner, we evaluate three candidate trajectories for each local target.

Figure \ref{fig:storyboard} depicts snapshots during a sample run of our framework using the Attentive Kernel for the terrain GP, in which the robot safely traverses the environment, improving its estimation of the terrain until it is able to reach the global goal by navigating around the obstacle regions. Figure \ref{fig:results} shows the final results for the same run, illustrating the accuracy of the learned terrain model. All simulations were run on a laptop with an Intel i7 CPU and 16 GB of RAM.

We also implement the simulation using full-order dynamics in MuJoCo, as depicted in Figure \ref{fig:highlight}. We use a variation of the angular momentum LIP planner~\cite{Gong2022AngularMomentum} to track the PSP plans as introduced in~\cite{shamsah2023integrated}. PSP hyperparameters (e.g., CoM velocities, and heading change) are used to design full-body joint trajectories through geometric inverse kinematics. A passivity-based controller~\cite{sadeghian2017passivity} is used for full-body trajectory tracking. The supplemental video for this work shows Digit navigating through multiple environments using our framework.\renewcommand{\thefootnote}{\roman{footnote}}\footnotemark[1]


\begin{table}[ht]
\centering
\caption{Benchmarking for Evaluated GPs}
\vspace{-5pt}
\begin{tabular}{|c |c |c |c|} 
 \hline
 Metric & Attentive Kernel & RBF Kernel & NN Kernel \\
 \hline
 \vspace{-5pt}
 Avg. Error (Path) & \thead{2.04e-4} & 1.10e-3 & 2.83e-4\\
 \vspace{-5pt}
  Avg. Std. Dev. (Path) & \thead{2.58e-2} & 2.67e-2 & 5.19e-2\\
 \vspace{-5pt}
 Avg. Error (Env.) & 52.60 & 49.61 & \thead{49.02} \\
 \vspace{-5pt}
  Avg. Std. Dev. (Env.) & \thead{3.13e-2} & 3.13e-2 & 5.48e-2\\
 \vspace{-5pt}
 Avg. Time [sec] & \thead{303} & 313.48 & 352.42\\
 \vspace{-5pt}
 Avg. Steps & 638 & \thead{617} & 667\\
 Retrain every \# steps & \thead{20} & 14 & 19.33\\
 \hline
\end{tabular}
\vspace{-0.1in}
\label{table:Benchmark}
\end{table}



In Table \ref{table:Benchmark}, we benchmark the efficacy of the Attentive Kernel, RBF, and NN Kernel GPs in our case studies. For each of the three environments, we run experiments using each GP until we achieve three successful runs. This is achieved by incrementally increasing the retraining frequency of the GP (measured in number of  walking steps taken by the robot). This iterative approach ensures that the GP model can provide accurate predictions of the local terrain elevation, thus enabling the bipedal robot to navigate locally without encountering impassable terrain heights. 
Overall, the Attentive kernel demonstrates the lowest prediction error along the traversed trajectory (Path), \textit{i.e.}, the path where the robot actually traversed, while the Neural Network kernel (with KD-tree and K-mean) excels in minimizing the prediction error across the global terrain map (Env.), \textit{i.e.}, the entire environment.
Furthermore, the Attentive kernel showcases notable proficiency in uncertainty quantification, evident in both the local traversed trajectory and the global terrain map, with the lowest average standard deviation for each prediction. Both the Attentive Kernel and the Neural Network kernel exhibit comparable retraining frequencies, allowing for a greater number of walking steps before updating the GP model. However, the Attentive kernel emerges as the more computationally efficient option, characterized by a lower average total trajectory time.
Lastly, although the RBF kernel facilitates the fewest total walking steps to reach the goal, it requires more frequent retraining of the GP model to achieve three successful runs.

\begin{figure}
    \centering
    \subfigure[]{\includegraphics[width=0.23\textwidth]{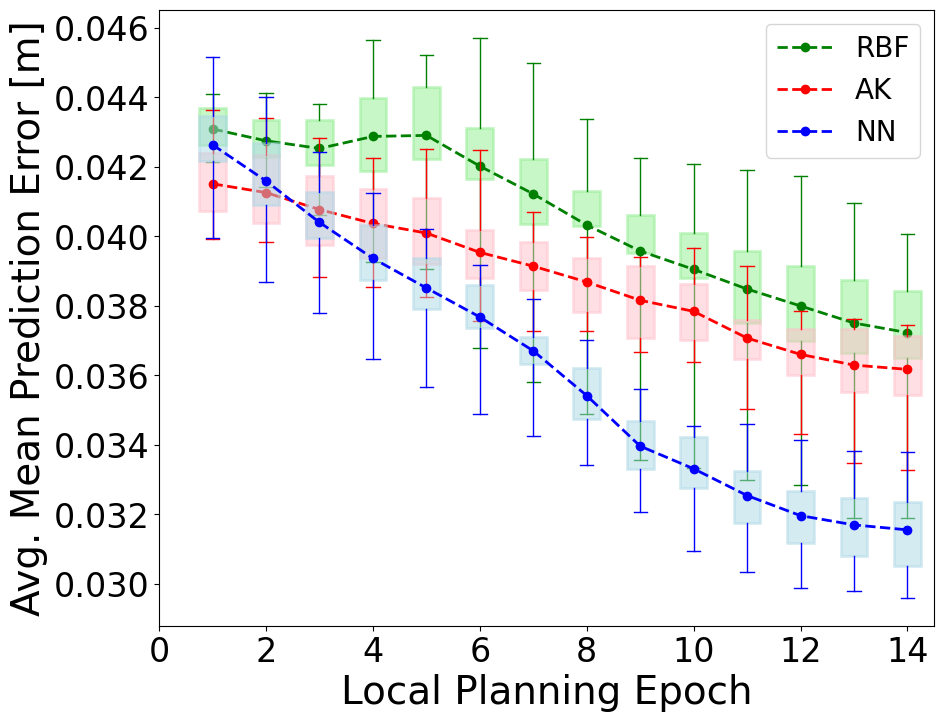}} 
    \subfigure[]{\includegraphics[width=0.23\textwidth]{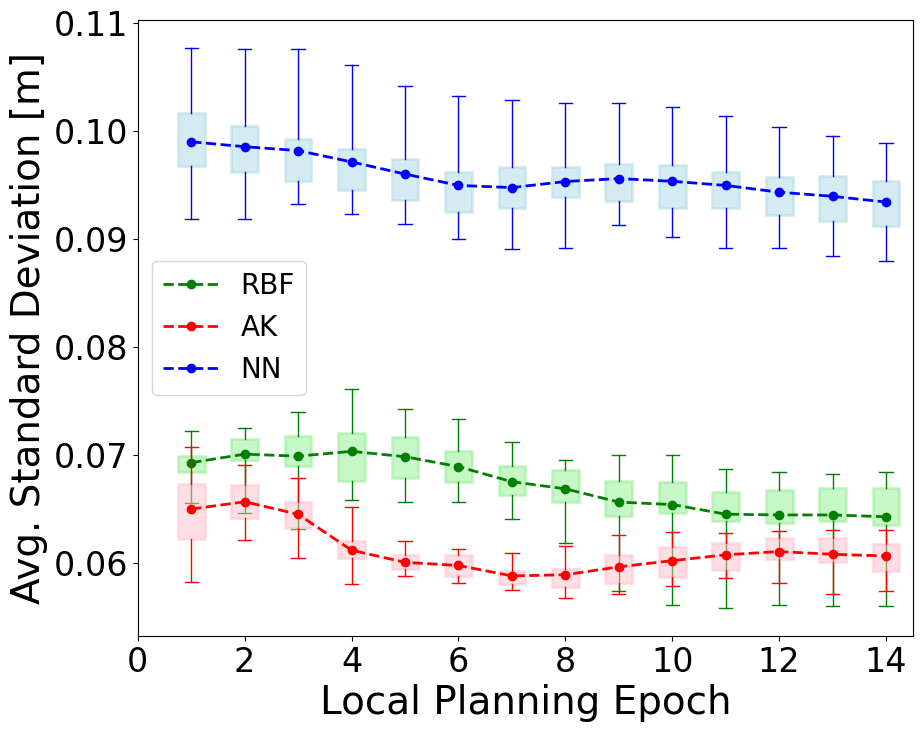}} 
    \subfigure[]{\includegraphics[width=0.23\textwidth]{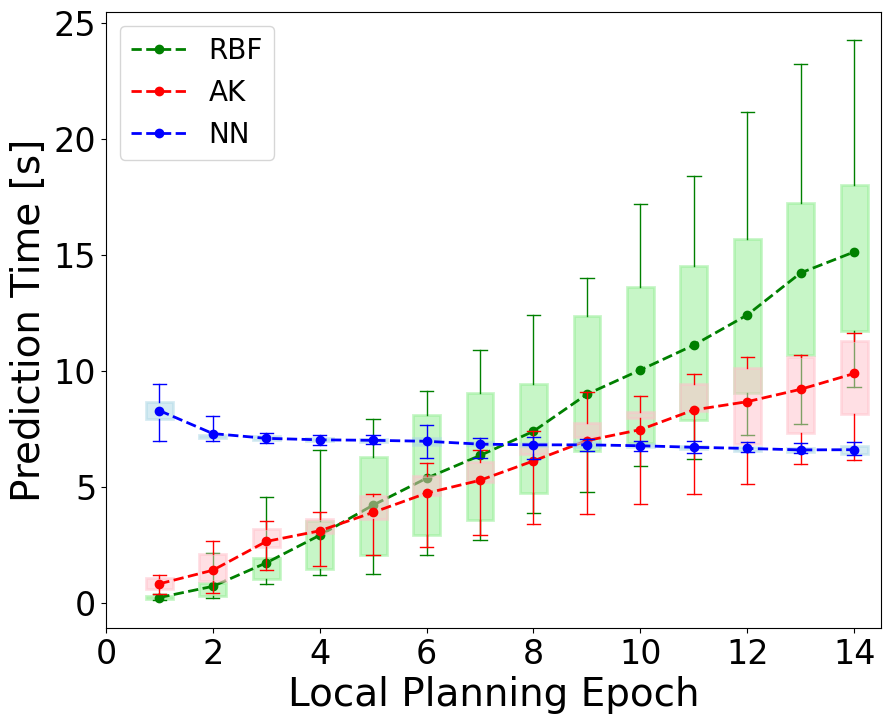}}
    \subfigure[]{\includegraphics[width=0.23\textwidth]{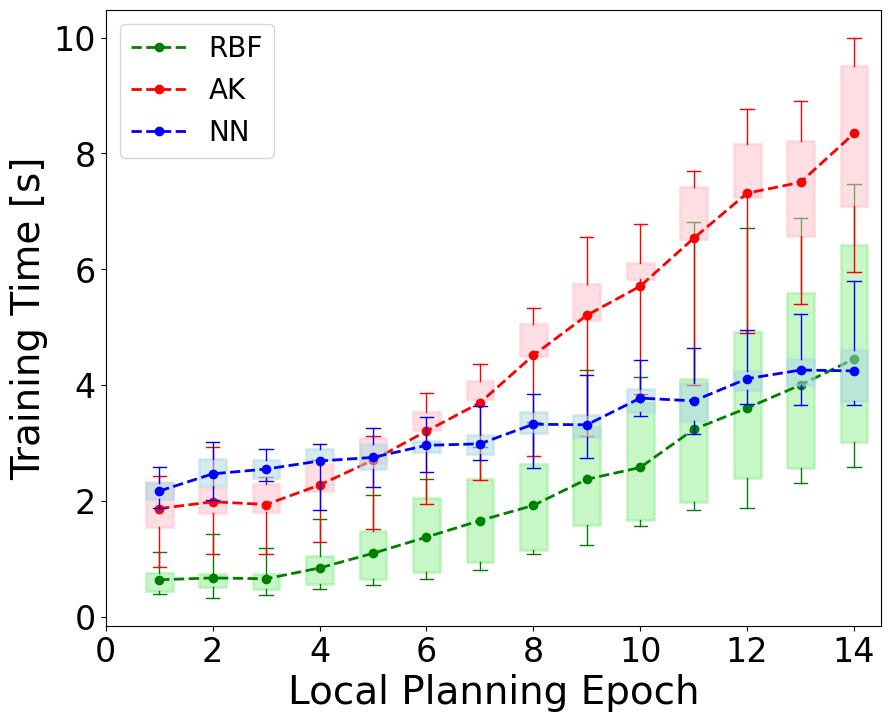}}
    \caption{Benchmarking of three considered GP kernels: (a) Average mean prediction error across the entire environment [(Ground Truth) - (GP estimated mean)] (b) Average standard deviation across the entire environment (c) Average GP prediction times (d) Average GP retraining times}
    \label{fig:Gp_comparative}
\end{figure}

In Figure \ref{fig:Gp_comparative}, we conducted additional experiments on terrain N37W112, averaging over 10 successful trials for each kernel. First, our findings reveal that the Neural Network kernel demonstrates the lowest mean prediction error for estimating the global terrain map, albeit with the highest uncertainty (average standard deviation) among the three kernels. Conversely, the Attentive Kernel exhibits superior performance in uncertainty quantification with the lowest average standard deviation throughout all local planning epochs, consistent with our previous benchmarking results, and displays enhanced prediction accuracy compared to the baseline RBF kernel.

With respect to computational complexity, the Neural Network kernel initially shows higher prediction times due to its utilization of a local approximation method, as discussed in \ref{subsection: GP Definition}. This method necessitates separate predictions for each cluster of query points and results in an extended duration for multiple predictions. However, the local approximation methods enable the Neural Network kernel to remain robust against the accumulation of newly-collected training data, as it consistently requires the same number of training datapoints for prediction. Consequently, the prediction time remains relatively consistent regardless of the total number of training datapoints amassed.
In contrast, both the Attentive kernel and the RBF kernel initially demonstrate rapid prediction computation but increase in prediction time as the number of training samples grows. This scaling arises from both kernels' utilization of all training data to predict terrain elevations, leading to their prediction times eventually exceeding those of the Neural Network kernel around the $9^{\rm th}$ local planning epoch.
Lastly, the training time for the Neural Network kernel and the RBF kernel is comparable, whereas the Attentive kernel necessitates additional training time for optimizing the hyperparameters for weighted summation of its various base kernels and for optimizing the membership vector assigned to each input location.

In summary, the Attentive kernel excels in uncertainty quantification and locally-traversed terrain elevation prediction, while the Neural Network kernel demonstrates better performance in predicting the entire environment despite exhibiting higher uncertainty. For long-horizon trajectory planning (or trajectory longer than our case study), the Neural Network kernel combined with a local approximation method is recommended for terrain GP modeling due to its consistent computational time for terrain estimation at every epoch. While the Attentive kernel can also be employed for long-horizon trajectories, the implementation of data allocation methods is advised to alleviate the computational complexity during the prediction phase. The same local approximation methods may be applied to the Attentive kernel, although the prediction accuracy and uncertainty quantification performance can potentially be compromised. Overall, both the Neural Network kernel and the Attentive kernel demonstrate superior performance compared to the baseline RBF kernel, attributed to their adaptability to the non-stationary nature of terrain.

\section{Conclusion}
In this work, we propose a novel hierarchical planning framework for bipedal navigation in rough and uncertain terrain environments using GP learning of uncertainty. Future work will implement this planning framework on hardware experiments of Digit navigating through outdoor fields.

\section*{Acknowledgment}
The authors would like to express our gratitude to Hyunyoung Jung for assistance in configuring the Digit simulation within the MuJoCo environment. Additionally, gratitude is extended to Weizhe Chen for providing the code enabling the visualization of the Gaussian Process estimated terrain elevation map and its associated uncertainty map.

\bibliographystyle{IEEEtran}
\bibliography{main}

\begin{thebibliography}{10}
\providecommand{\url}[1]{#1}
\csname url@samestyle\endcsname
\providecommand{\newblock}{\relax}
\providecommand{\bibinfo}[2]{#2}
\providecommand{\BIBentrySTDinterwordspacing}{\spaceskip=0pt\relax}
\providecommand{\BIBentryALTinterwordstretchfactor}{4}
\providecommand{\BIBentryALTinterwordspacing}{\spaceskip=\fontdimen2\font plus
\BIBentryALTinterwordstretchfactor\fontdimen3\font minus \fontdimen4\font\relax}
\providecommand{\BIBforeignlanguage}[2]{{%
\expandafter\ifx\csname l@#1\endcsname\relax
\typeout{** WARNING: IEEEtran.bst: No hyphenation pattern has been}%
\typeout{** loaded for the language `#1'. Using the pattern for}%
\typeout{** the default language instead.}%
\else
\language=\csname l@#1\endcsname
\fi
#2}}
\providecommand{\BIBdecl}{\relax}
\BIBdecl

\bibitem{torres2022legged}
A.~Torres-Pardo, D.~Pinto-Fern{\'a}ndez, M.~Garabini, F.~Angelini, D.~Rodriguez-Cianca, S.~Massardi, J.~Tornero, J.~C. Moreno, and D.~Torricelli, ``Legged locomotion over irregular terrains: State of the art of human and robot performance,'' \emph{Bioinspiration \& Biomimetics}, vol.~17, no.~6, p. 061002, 2022.

\bibitem{gibson2022terrain}
G.~Gibson, O.~Dosunmu-Ogunbi, Y.~Gong, and J.~Grizzle, ``Terrain-adaptive, alip-based bipedal locomotion controller via model predictive control and virtual constraints,'' in \emph{IEEE/RSJ International Conference on Intelligent Robots and Systems}.\hskip 1em plus 0.5em minus 0.4em\relax IEEE, 2022, pp. 6724--6731.

\bibitem{huang2023efficient}
J.-K. Huang and J.~W. Grizzle, ``Efficient anytime clf reactive planning system for a bipedal robot on undulating terrain,'' \emph{IEEE Transactions on Robotics}, 2023.

\bibitem{DataS2S_Dai}
M.~Dai, X.~Xiong, and A.~D. Ames, ``Data-driven step-to-step dynamics based adaptive control for robust and versatile underactuated bipedal robotic walking,'' 2022.

\bibitem{krishna2022linear}
L.~Krishna, G.~A. Castillo, U.~A. Mishra, A.~Hereid, and S.~Kolathaya, ``Linear policies are sufficient to realize robust bipedal walking on challenging terrains,'' \emph{IEEE Robotics and Automation Letters}, vol.~7, no.~2, pp. 2047--2054, 2022.

\bibitem{wu2023infer}
F.~Wu, Z.~Gu, H.~Wu, A.~Wu, and Y.~Zhao, ``Infer and adapt: Bipedal locomotion reward learning from demonstrations via inverse reinforcement learning,'' in \emph{IEEE International Conference on Robotics and Automation}, 2024.

\bibitem{yang2013gaussian}
K.~Yang, S.~Keat~Gan, and S.~Sukkarieh, ``A gaussian process-based rrt planner for the exploration of an unknown and cluttered environment with a uav,'' \emph{Advanced Robotics}, vol.~27, no.~6, pp. 431--443, 2013.

\bibitem{barbosa2021risk}
F.~S. Barbosa, B.~Lacerda, P.~Duckworth, J.~Tumova, and N.~Hawes, ``Risk-aware motion planning in partially known environments,'' in \emph{IEEE Conference on Decision and Control}, 2021, pp. 5220--5226.

\bibitem{viseras2019robotic}
A.~Viseras, D.~Shutin, and L.~Merino, ``Robotic active information gathering for spatial field reconstruction with rapidly-exploring random trees and online learning of gaussian processes,'' \emph{Sensors}, vol.~19, no.~5, p. 1016, 2019.

\bibitem{jian2023path}
Z.~Jian, Z.~Liu, H.~Shao, X.~Wang, X.~Chen, and B.~Liang, ``Path generation for wheeled robots autonomous navigation on vegetated terrain,'' \emph{IEEE Robotics and Automation Letters}, 2023.

\bibitem{kanoulas2018footstep}
D.~Kanoulas, A.~Stumpf, V.~S. Raghavan, C.~Zhou, A.~Toumpa, O.~Von~Stryk, D.~G. Caldwell, and N.~G. Tsagarakis, ``Footstep planning in rough terrain for bipedal robots using curved contact patches,'' in \emph{2018 IEEE International Conference on Robotics and Automation (ICRA)}, 2018, pp. 4662--4669.

\bibitem{bertrand2020detecting}
S.~Bertrand, I.~Lee, B.~Mishra, D.~Calvert, J.~Pratt, and R.~Griffin, ``Detecting usable planar regions for legged robot locomotion,'' in \emph{2020 IEEE/RSJ International Conference on Intelligent Robots and Systems (IROS)}, 2020, pp. 4736--4742.

\bibitem{chen2022ak}
W.~Chen, R.~Khardon, and L.~Liu, ``Ak: Attentive kernel for information gathering,'' in \emph{Robotics: Science and Systems (RSS)}, 2022.

\bibitem{vasudevan2009gaussian}
S.~Vasudevan, F.~Ramos, E.~Nettleton, H.~Durrant-Whyte, and A.~Blair, ``Gaussian process modeling of large scale terrain,'' in \emph{IEEE International Conference on Robotics and Automation}, 2009, pp. 1047--1053.

\bibitem{plagemann2009bayesian}
C.~Plagemann, S.~Mischke, S.~Prentice, K.~Kersting, N.~Roy, and W.~Burgard, ``A bayesian regression approach to terrain mapping and an application to legged robot locomotion,'' \emph{Journal of Field Robotics}, vol.~26, no.~10, pp. 789--811, 2009.

\bibitem{seyde2019locmotion}
T.~Seyde, J.~Carius, R.~Grandia, F.~Farshidian, and M.~Hutter, ``Locomotion planning through a hybrid bayesian trajectory optimization,'' in \emph{International Conference on Robotics and Automation}, 2019, pp. 5544--5550.

\bibitem{homberger2019support}
T.~Homberger, L.~Wellhausen, P.~Fankhauser, and M.~Hutter, ``Support surface estimation for legged robots,'' in \emph{International Conference on Robotics and Automation}.\hskip 1em plus 0.5em minus 0.4em\relax IEEE, 2019, pp. 8470--8476.

\bibitem{todorov2012mujoco}
E.~Todorov, T.~Erez, and Y.~Tassa, ``Mujoco: A physics engine for model-based control,'' in \emph{2012 IEEE/RSJ International Conference on Intelligent Robots and Systems}, 2012, pp. 5026--5033.

\bibitem{zhao2017robust}
Y.~Zhao, B.~R. Fernandez, and L.~Sentis, ``Robust optimal planning and control of non-periodic bipedal locomotion with a centroidal momentum model,'' \emph{The International Journal of Robotics Research}, vol.~36, no.~11, pp. 1211--1242, 2017.

\bibitem{shamsah2023integrated}
A.~Shamsah, Z.~Gu, J.~Warnke, S.~Hutchinson, and Y.~Zhao, ``Integrated task and motion planning for safe legged navigation in partially observable environments,'' \emph{IEEE Transactions on Robotics}, pp. 1--22, 2023.

\bibitem{leibfried2021tutorial}
F.~Leibfried, V.~Dutordoir, S.~John, and N.~Durrande, ``A tutorial on sparse gaussian processes and variational inference,'' 2021, arXiv: 2012.13962 [cs.LG].

\bibitem{neal2012bayesian}
R.~M. Neal, \emph{Bayesian learning for neural networks}.\hskip 1em plus 0.5em minus 0.4em\relax Springer Science \& Business Media, 2012, vol. 118.

\bibitem{jiang2023abstraction}
J.~Jiang, S.~Coogan, and Y.~Zhao, ``Abstraction-based planning for uncertainty-aware legged navigation,'' \emph{IEEE Open Journal of Control Systems}, 2023.

\bibitem{Gong2022AngularMomentum}
Y.~Gong and J.~W. Grizzle, ``{Zero Dynamics, Pendulum Models, and Angular Momentum in Feedback Control of Bipedal Locomotion},'' \emph{Journal of Dynamic Systems, Measurement, and Control}, vol. 144, no.~12, 10 2022, 121006.

\bibitem{sadeghian2017passivity}
H.~Sadeghian, C.~Ott, G.~Garofalo, and G.~Cheng, ``Passivity-based control of underactuated biped robots within hybrid zero dynamics approach,'' in \emph{IEEE International Conference on Robotics and Automation}.\hskip 1em plus 0.5em minus 0.4em\relax IEEE, 2017, pp. 4096--4101.

\end{thebibliography}

\end{document}